\documentclass[runningheads]{llncs}
\usepackage{graphicx}
\usepackage{times}
\usepackage{epsfig}
\usepackage{amsmath}
\usepackage{amssymb}
\usepackage{booktabs}
\usepackage{xcolor}
\usepackage{wrapfig}
\usepackage{orcidlink}

\begin{document}
\title{Graphing the Future: Activity and Next Active Object Prediction using Graph-based Activity Representations}
\titlerunning{Graphing the Future}
%
%
\author{Victoria Manousaki\inst{1,2}\orcidlink{0000-0003-2181-791X} \and
Konstantinos Papoutsakis\inst{2}\orcidlink{0000-0002-2467-8727} 
\and
Antonis Argyros\inst{1,2}\orcidlink{0000-0001-8230-3192} }

\institute{Computer Science Department, University of Crete \and
Institute of Computer Science, Foundation for Research and Technology - Hellas (FORTH)
\email{}\\
\email{\{vmanous,papoutsa,argyros\}@ics.forth.gr}
}
\setcounter{footnote}{0}
%
%
\maketitle              
\begin{abstract}
We present a novel approach for the visual prediction of human-object interactions in videos. Rather than forecasting the human and object motion or the future hand-object contact points, we aim at predicting (a)~the class of the on-going human-object interaction and (b)~the class(es) of the next active object(s) (NAOs), i.e., the object(s) that will be involved in the interaction in the near future as well as the time the interaction will occur. Graph matching relies on the efficient Graph Edit distance (GED) method. The experimental evaluation of the proposed approach was conducted using two well-established video datasets that contain human-object interactions, namely the MSR Daily Activities and the CAD120. High prediction accuracy was obtained for both action prediction and NAO forecasting.

\end{abstract}
\section{Introduction}

Prediction provides smart agents the ability to take a look into the future in order to proactively foresee possible outcomes or adverse, high-risk events. This enables them to plan timely responses for early intervention or corrective actions~\cite{hu2022online,kong2022human,oprea2020review}.
Such a competence is rather important when it comes to the observation of the environment or scenes in a wide variety of applications such as assistive robots in domestic or industrial environments~\cite{petkovic2022human} or pedestrian/obstacle trajectory prediction for autonomous vehicles~\cite{wang2022stepwise} and more. 
Our study focuses on prediction of the semantics of a partially observed activity, before its completion, and of the next active objects that will be involved in order to complete the ongoing activity.
Specifically, the proposed approach aspires to model the spatio-temporal relationships between the human and the visible scene objects in order to predict the classes of a varying number of the next active objects that will be handled by the human in order to complete the ongoing activity. Current methods lack the ability to predict more than one next active object~\cite{dessalene2021forecasting,furnari2017next,furnari2020rolling}. To the best of our knowledge, this is the first approach that is able to jointly predict the semantics of the ongoing activity and multiple next active objects.  
Moreover, one aspect that can be of great importance to such prediction systems is the ability to forecast the time in which NAOs will be involved in the current scenario. Our method is the first to predict NAOs along with the time that they will be involved in the activity.

\begin{figure}
\centering
\includegraphics[width=.5\textwidth]{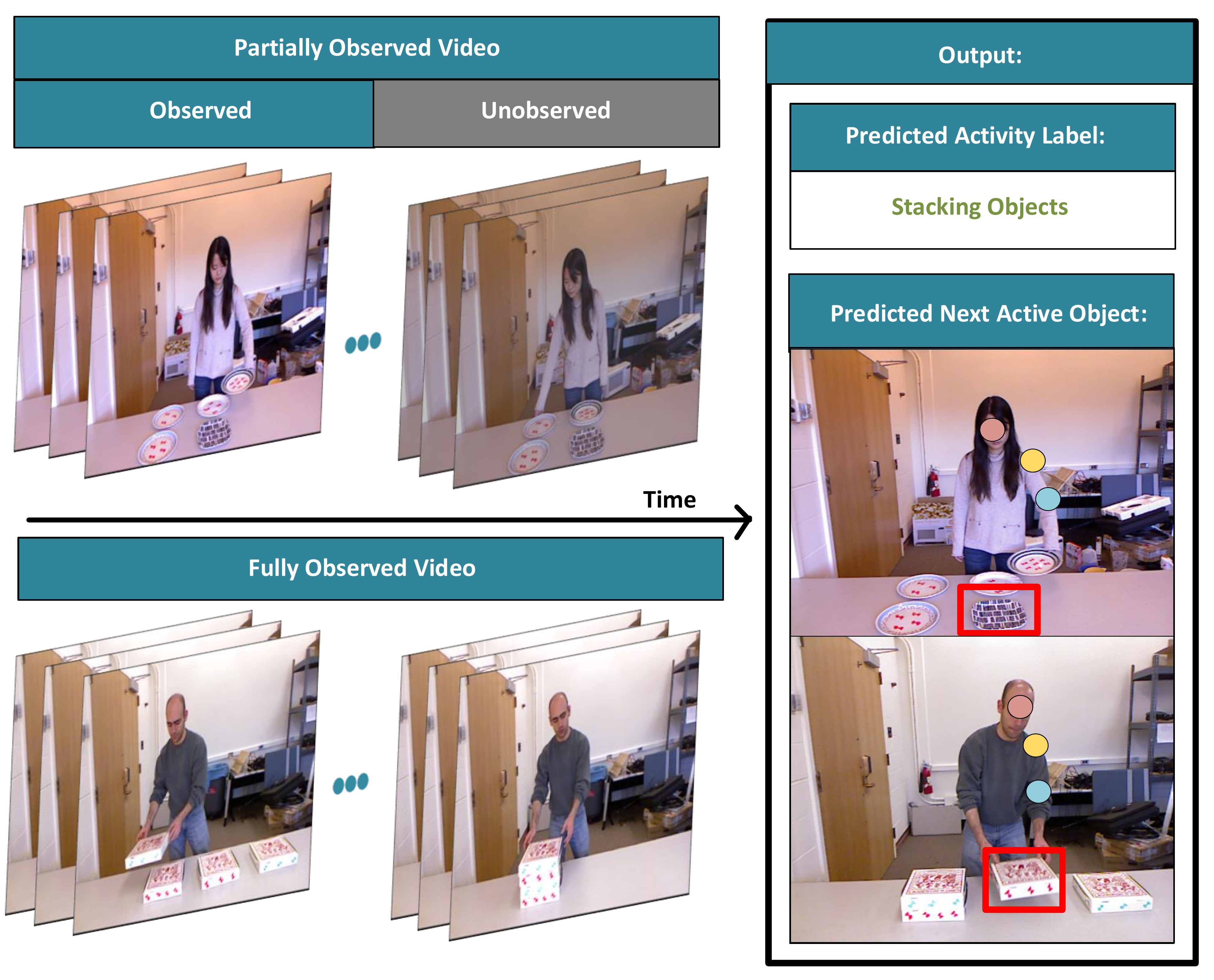} 
\caption{By matching a partially executed and observed activity, to a prototype, fully observed one, we are able to infer correspondences of similar objects and human joints between the two videos. This, in turn, enables to perform activity and next-active-object prediction in the partially observed activity. The example in this figure refers to the ``stacking objects'' activity, which is performed with a different number and types of objects in the partially and the fully observed activities.
}
\label{fig:matches}
\end{figure}

In this paper, we propose to jointly forecast the activity and the objects that will participate in the execution of the activity till its completion. Instead of predicting the interaction hotspots~\cite{liu2022joint,liu2020forecasting,nagarajan2019grounded} of a NAO,  we propose a holistic understanding of the activity regarding the human and objects present in the scene. Our approach is based upon calculating the dissimilarity of graphs representing the entities that constitute the activity~\cite{papoutsakis2019unsupervised}. 
Specifically, the human body joints of the acting person and  the scene objects are represented as nodes of a graph and the semantic and motion relations between the nodes are represented as edges. The dissimilarity of graphs is calculated using the graph edit distance (GED)~\cite{abuaisheh}.

We showcase our approach on video datasets of  human-object interactions of varying complexity. The well-known MSR-Daily Activities dataset~\cite{wang2012mining} includes activities where none or one object is handled by a single subject. 
We further evaluate the performance of the proposed method using the CAD-120 dataset~\cite{koppula2013learning} that contains long and complex activities. Instances of the activities are performed by different subjects using different types and a varying number of objects. As an example, different executions of the ``stacking objects" activity are performed using 4 boxes and 5 plates, respectively (see Fig.~\ref{fig:matches}).
The main contributions can be summarized as follows:
\begin{itemize}
    \item We propose Graphing The Future (GTF), 
    a method that can jointly predict the activity label and the next-active-objects by calculating the dissimilarity of videos with the use of GED as well as the time instance at which these objects will be used in the ongoing activity.
    \item Our work is the first to address the prediction of multiple NAOs in human-object interaction scenarios.

    \item GTF models the pairwise correspondences of objects and human joints between two comparing videos based on their semantic similarity as well as their (intra-video) spatio-temporal relationships in each video. Therefore, predictions are in principle possible even when a particular interaction with an object of a specific class has never been observed before.

\end{itemize}

\vspace{-4mm}
\section{Related Work}
\label{sec::related}
\vspace{-2mm}
\vspace*{0.2cm}\noindent{\textbf{Activity Prediction:}} Action prediction aims to forecast the label of an action based on limited/partial observations. The majority of the proposed methods that tackle this problem consider (first person) egocentric videos~\cite{rodin2021predicting,zatsarynna2021multi,xu2022learning,sener2020temporal,abu2020long}, mainly due to the availability of large amounts of relevant video data and annotations~\cite{grauman2022ego4d,damen2022rescaling,ragusa2021meccano}. In~\cite{girdhar2021anticipative}, Video Transformers are proposed to accurately anticipate future actions. Without supervision the method learns to focus on the image areas where the hands and objects appear, while attends the most relevant frames for the prediction of the next action. 
Rodin et al.~\cite{rodin2022untrimmed} tackles the problem of anticipation in untrimmed videos in an attempt to generalize and deal with unconstrained conditions in real world scenarios. 
An advantage of the work proposed by Furnari et al.~\cite{furnari2019would,furnari2020rolling} is the ability to make predictions not only in first-person but also in third-person videos. Their work focuses on making predictions using multiple modalities such as RGB frames, optical flow and object-based features. Their architecture uses one LSTM for encoding the past time steps while the second LSTM makes predictions about the future. Manousaki et al.~\cite{manousaki2021action,manousaki2022segregational} focused their work on predicting action sequences by using temporal alignment algorithms. They aligned complete and partially observed actions using the Segregational Soft Dynamic Time Warping (SSDTW) algorithm by fusing the human and object motion. Wu et al.~\cite{wu2021spatial} opted to solve the problem of activity prediction by exploring spatio-temporal relations between humans and objects. They used a graph-based neural network to encode the spatial relations between video entities at different time-scales.

\begin{figure}[t]
\centering
\includegraphics[width=\textwidth]{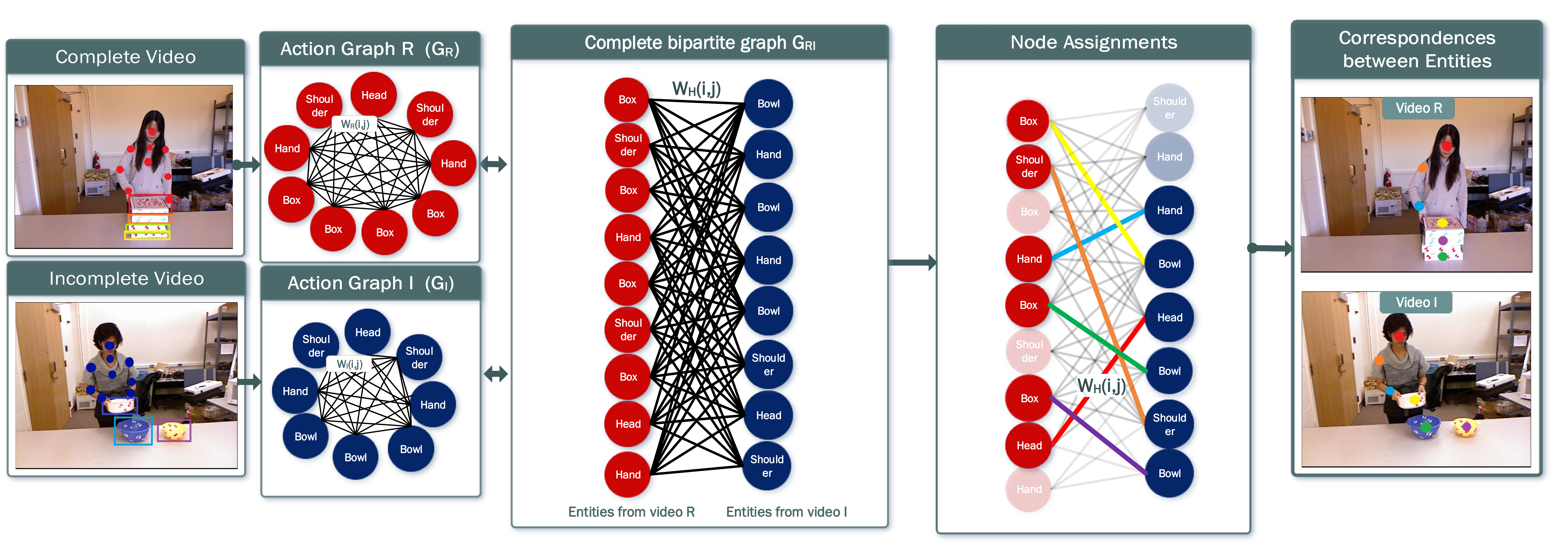} 
\caption{Graph matching of a complete video (reference) and an incomplete/partially observed (test) video. First, the fully connected graphs of each video are created based on the video entities. On the basis of these graphs, a bipartite graph between the action graphs is constructed. By calculating the GED, we are able to correspond nodes  between the two original action graphs.
}
\label{fig:concept}
\end{figure}

\vspace*{0.04cm}\noindent{\textbf{Next-Active-Object Prediction:}} Having correctly predicted the activity label, recent studies focus their attention on predicting the next-active-object. 
Dessalene et al.~\cite{dessalene2021forecasting} define an active object as the object presently \textit{in contact with a hand} while next-active-object is the object which \textit{will next come into contact with that hand}. We argue that an object can be the next-active-object without having the need to come in contact with the hand. For example, imagine a scenario in which a hand pushes an object, which comes in contact with another object which is pushed, too. The hand never comes in contact with the second object. However, the second object is definitely part of the interaction. So, we define next-active-object as the object that is the next to be involved in the progress of an action. 

In the course of an activity many actions can take place. These actions can be performed with or without the use of objects. Some consecutive actions may use the same object. In case there is no change of the active object between actions, the object used in consequent actions is not considered as next-active-object only because the action has changed. 
Our work differs from other approaches towards the prediction of objects. Other approaches~\cite{girdhar2021anticipative,zatsarynna2021multi} perform prediction of the object of the next segment/action, which in some cases can be the current active object of the ongoing segment. Liu et al.~\cite{liu2022joint} predict future hand trajectories and object interaction hotspots, while in~\cite{yagi2021hand} hand-object contact prediction (contact or no-contact) is modelled using hand and object tracks throughout the video. 
This task if different from our target task of next-active-object prediction.

The first approach to tackle the problem of next-active-object prediction was Furnari et al.~\cite{furnari2017next}. A sliding window was utilized in conjunction with an object detector in order to model each tracked trajectory and classify it as passive or active using random forests. The paper argues that the next-active-object can be distinguished from its frames immediately before it turns active. One very interesting characteristic of the method they propose is its ability to generalise to unseen object classes. However, their experiments show a loss of accuracy when dealing with unseen object classes thus proposing to train the method with the object classes that will be present in the test set for better results.

The work of Dessalene et al.~\cite{dessalene2021forecasting} employs graphs to predict the partially observed action and produce Contact Anticipation Maps which provide pixel-wise information of the anticipated time-to-contact involving one hand, either the left or the right. Also, they perform next-active-object segmentation by localizing candidate next active objects. These localizations are evaluated with the calculation of the Intersection over Union (IoU) value of the bounding boxes produced from the Faster-RCNN model. This work predicts the hand-object time-to-contact in egocentric videos but this does imply that this can be the next-active-object or that this object will be used immediately. Also, this is trained on annotated object classes of the dataset which implies that it cannot generalize to unseen object classes.

\section{The Proposed Method - GTF}
\label{sec::methodology}
We introduce the GTF method that jointly tackles the tasks of activity prediction and of next active object(s) prediction in videos using graph-based representation of an activity and graph matching technique based on the Graph Edit Distance measure to compare pairs of videos.
The \textit{activity prediction} task can be defined as the problem of inferring the label of an ongoing activity before its actual completion. Let an activity, noted as $A$, that starts at time $t_s$ and ends at time $t_e$, thus has a duration $d=t_e-t_s$. Its observation time is defined in proportions of $10\%$ of $d$. The goal is to predict the correct class as early as possible which implies access to fewer observations. 
We also note the task of \textit{next-active-object prediction} as the problem of the inference of the semantic label of an object that will be used in the progress of an activity. Multiple objects may be used in the progress of a given activity $A$. Related works~\cite{dessalene2021forecasting,furnari2020rolling} predict the next-active-object in the segment preceding it's use, i.e., an amount of time (measured in seconds) before the start of the action that involves the object of interest.


\vspace*{0.04cm}
Our approach relies on a graph-based representation of an activity that is captured in video. The entities in a video regard the tracked human skeletal joints and the observable/visible objects. Each video entity is represented as a node of an undirected graph, which also models both semantic information (object label) and its motion (2D or 3D trajectory). Each graph edge connecting two nodes represents the semantic similarity and the spatio-temporal relationships of the interconnected video entities, as described in Section~\ref{ssec:semantic}.
Our goal is to devise a novel approach that is able to identify human joints and/or objects in two different videos, one fully and one partially observed video, that exhibit similar behaviors and interactions with other entities using bipartite graph-matching. As shown in Fig.~\ref{fig:concept}\, a fully and a partially observed video are represented as two action graphs whose nodes represent the detected and tracked objects and human joints.

\vspace*{0.04cm} \noindent{\textbf{{Video Representation:}}}
Given a video of duration $T$ frames, it can be seen, at an object-level, as a complete and undirected graph, noted as $G = (V,E)$. 
In the course of a video, entities such as human body joints and foreground objects are localized and tracked using 2D or 3D human body pose estimation and tracking as well as object detection methods, respectively.
Each graph node is noted as $v \in V$ and graph edges are noted as $e_{ij} = (v_i , v_j) \in E$ between nodes $v_i, v_j \in V$, where $i \neq j$. The relations between the nodes describe their dissimilarity in the form of edge weights. The dissimilarity is described based on the semantic dissimilarity $s_i$ and the motion dissimilarity $m_i$. The edge weight between two connected nodes is defined as the weighted sum of the semantic and motion dissimilarity as follows:
\begin{equation}
    w_{ij} = (1-\lambda) * m_{ij} + \lambda * s_{ij}.
\label{eq::weights}
\end{equation}
The parameter $lamda \in [0,1]$ is user-defined and controls the  contribution of the semantic and motion information. On the extremity of $lamda = 0$, only motion information is considered while when $lamda = 1$, only semantic information is used. In the experimental section of this paper, we present an investigation of the effect of this parameter on the performance of the proposed method. 

\vspace*{0.04cm} \noindent{\textbf{{Semantic Dissimilarity:}}}
\label{ssec:semantic}
The weights $s_{ij}$  represent the semantic dissimilarity between the labels of the nodes $v_i$ and $v_j$. The node labels are retrieved based on ground truth annotations or object recognition methods. The semantic similarity of nodes $v_i$ and $v_j$ with recognized labels $l_i$ and $l_j$ is described as $S(l_i,l_j)$ and is estimated using the WordNet~\cite{fellbaum2005wordnet} lexical database and the Natural Language Toolkit~\cite{loper2002nltk} to compute the path-based Wu-Palmer scaled metric~\cite{wu1994verb}. The similarity is in the range $(0,1]$ with $1$ identifying identical words so semantic weight is:
\begin{equation}
    s_{ij} = 1 - S(l_i ,l_j).   
\end{equation}

\vspace*{0.04cm} \noindent{\textbf{{Motion Dissimilarity:}}}
\label{ssec:motion}
Each node in the graph is described by a feature vector which can encode information such as the 2D/3D human joint location, the 2D/3D location of the object centroid or any other feature such as appearance, optical flow, etc. The extracted motion features for each dataset are described in section~\ref{ssec::feat_extraction}. The acquired 2D/3D skeletal-based pose features or the 2D/3D object-based pose features are described by a trajectory $t(v_i)$ encoding the movement of the video entity during the activity. A pair of trajectories $t(v_i)$ and $t(v_j)$ can be aligned temporally using the Segregational Soft Dynamic Time Warping (SSDTW)~\cite{manousaki2022segregational} algorithm. The alignment cost of the trajectories $t(v_i)$ and $t(v_j)$ describes the motion dissimilarity of the graph nodes $v_i$ and $v_j$ and is divided by the summation of the length of the trajectory of the incomplete sequence $t(v_i)$ and the length of the trajectory of the reference sequence $t(v_j)$ that matched with $t(v_i)$ as proposed by the authors~\cite{manousaki2022segregational}. Thus, the weight $m_{ij}$ of an edge connecting the graph nodes $v_i$ and $v_j$ is:

\begin{equation}
    m_{i,j} = \frac{SSDTW(t(v_i) ,t(v_j))}{(len(t(v_i)) + len(t(v_j)))}.   
\end{equation}

\vspace*{0.04cm} \noindent{\textbf{{Graph Operations:}}}  
Having represented one partially observed and one complete video as graphs, we estimate their dissimilarity by using Graph Edit Distance (GED)~\cite{abuaisheh}. GED is calculated by considering the edit operations (insertions, deletions and substitutions of nodes and/or edges) that are needed in order to transform one graph into another with minimum cost. 
Our GTF approach is inspired by the approach of Papoutsakis et al.~\cite{papoutsakis2019unsupervised} which uses the GED in order to solve the problem of co-segmentation in triplets of videos. Different from~\cite{papoutsakis2019unsupervised} we propose to assess the GED between a pair of videos in order to perform activity prediction. Comparably to~\cite{papoutsakis2019unsupervised} our approach is based on semantic and motion similarity of the entities but instead of using the EVACO cosegmentation method~\cite{papoutsakis2017temporal} to compute the alignment cost of the co-segmented sub-sequences we employ the SSDTW algorithm~\cite{manousaki2022segregational} to align the trajectories between pairs of nodes. The SSDTW algorithm has been shown to have better performance in aligning incomplete/ partially observed sequences for the task of action prediction.

We create a graph for each video $G_I$ ((I)ncomplete video) and $G_R$ ((R)eference video) and assess their graph distance. $W_I$ and $W_R$ are the dissimilarity matrices of action graphs $G_I$ and $G_R$ with size $N_I  \times N_I$ and $N_R  \times N_R$, respectively, where $N_I$ and $N_R$ are the number of vertices of each graph. As seen in Fig.~\ref{fig:concept} the next step is to create the bipartite graph $G_{IR}$ of the action graphs $G_I$ and $G_R$. The edge weights $W_H$ connecting the nodes of graph $G_I$ to nodes of graph $G_R$ are calculated using Equation~(\ref{eq::weights}). In order to calculate the GED on the bipartite graph we need to employ the Bipartite Graph Edit Distance (BP-GED) which solves an assignment problem on the complete bipartite graph using the Kuhn-Munkres algorithm~\cite{munkres1957algorithms}. The weights of the complete bipartite graph $G_{IR}$ are: $W_{IR} =$ $\begin{bmatrix}
0_{N_I,N_I} & W_H \\
{W_H}^T & 0_{N_R,N_R},
\end{bmatrix}$
where $0_{x,y}$ stands for an $x \times y$ matrix of zeros. The solution of this assignment problem requires the definition of the graph edit operations and their associated costs.

\vspace*{0.04cm} \noindent{\textbf{{Node operations:}}} Consist of node insertions, deletions and substitutions. The cost of inserting and deleting a node $v$ is:

\begin{equation}
nd_{in}( empty\_node \xrightarrow[]{} v_i) = \tau_v,    \quad  
nd_{del}( v_i\xrightarrow[]{} empty\_node ) = \tau_v     
\end{equation}

while the cost of substitution of node $v$ with node $u$ is:
\begin{equation}
nd_{sb}( v_i \xrightarrow[]{} u_j ) = [\frac{1}{2\tau_v} + \exp{(-a_v * W_H(i,j) + \sigma_v)}]^{-1}  . 
\end{equation}
The parameters of the cost operations for the nodes where set experimentally to $\tau_v = 0.4$, $\alpha_v = 0.1$ and $\sigma_v = 0.0$.   

\vspace*{0.04cm} \noindent{\textbf{{Edge operations:}}}  also consist of insertions, deletions and substitutions. The costs of inserting and deleting an edge from node $n$ of graph $G_I$ to node $u$ of graph $G_R$ is:
\begin{equation}
e_{in}( e_{ij}^{G_I} \xrightarrow[]{} e_{mn}^{G_R}) = \tau_e,    \quad  
e_{del}( e_{ij}^{G_I} \xrightarrow[]{} e_{mn}^{G_R}) = \tau_e.
\end{equation}

Finally, the cost of edge substitution is defined as:
\begin{equation}
e_{sb}( e_{ij}^{G_I} \xrightarrow[]{} e_{mn}^{G_R}) = 
\left[ \frac{1}{2\tau_e} +  exp\left(-\alpha_e \cdot (W_I(i,j)+W_R(m,n))/2 + \sigma_e\right)\right]^{-1}.  
\end{equation}

The parameters of the cost operations for the edges where set experimentally to $\tau_e = 0.3$, $\alpha_e = 0.1$ and $\sigma_e = 100$.  

\vspace*{0.04cm} \noindent{\textbf{{Action distance:}}}
The dissimilarity between a pair of graphs ($G_I,G_R$) is computed by the BP-GED which calculates the exact GED~\cite{abuaisheh}. With GED the minimum edit operations are calculated for transforming graph $G_I$ to graph $G_R$. The dissimilarity, denoted as BP-GED($G_I,G_R$), in the work of~\cite{papoutsakis2019unsupervised} is normalized by the total number of objects. This normalization is effective when looking for commonalities between videos but is ineffective for activity prediction. In our work we need to be flexible in the number of objects that can be used during an activity while discarding irrelevant objects. In order to achieve this, we found that the best option is to normalize by the number of pairs of matched objects (MO). This helps us to assess our method on the objects that are important for the prediction and discard objects that may be present but with no use in the activity performed. Thus, the dissimilarity $D(G_I,G_R)$ of  graphs $G_I$, $G_R$ is defined as:
\begin{equation}
    D(G_I,G_R) = BP{\text -}GED(G_I,G_R)/MO.   
\end{equation}

\begin{figure}[t]
\centering
\includegraphics[width=.49\textwidth]{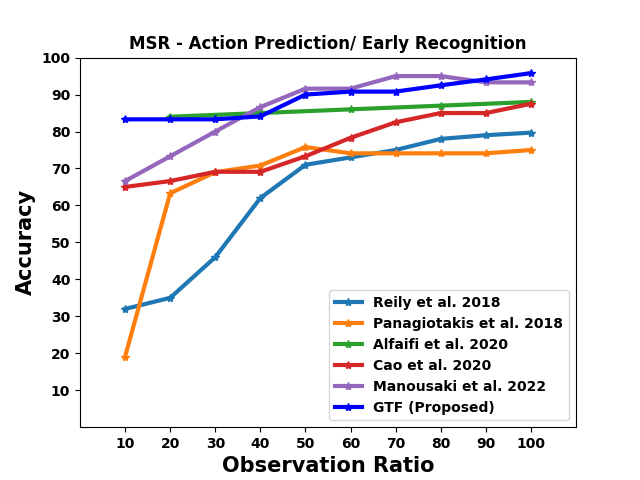} 
\includegraphics[width=.49\textwidth]{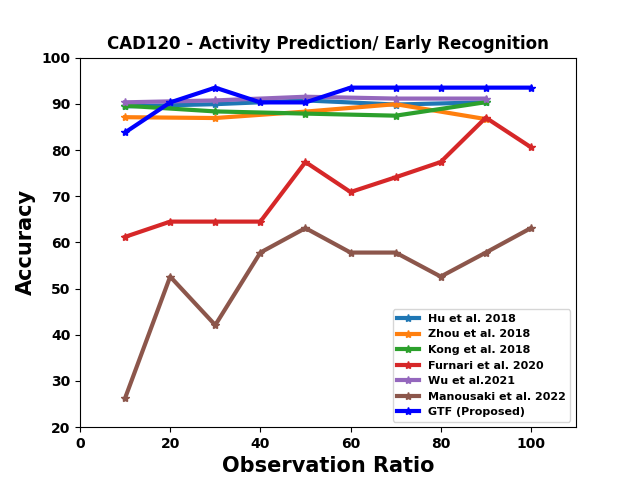} 
\caption{Activity prediction results for the (left) MSR Daily Activities and (right) CAD-120 datasets for different observation ratios.}
\label{fig:msr}
\end{figure}

\section{Experiments}
\label{sec::exp}
\subsection{Datasets}
\label{ssec::datasets}

\vspace*{0.04cm}\noindent \textit{{MSR Daily Activity 3D Dataset~\cite{wang2012mining}:}}
The activities contained in this dataset involve human-object interactions in trimmed video executions. The dataset contains 16 activity classes the executions of which are performed by male and female subjects, the first time by standing up and the second by laying down. The dataset contains the 3D locations of the human body joints. The evaluation split of the related works~\cite{reily2018skeleton,manousaki2021action,manousaki2022segregational} is used for a fair comparative evaluation.

\vspace*{0.04cm}\noindent \textit{{CAD-120 Dataset~\cite{koppula2013learning}:}}
Contains complex activities that represent human-object interactions performed by different subjects. The activities are performed using 10 different objects and are observed from varying viewpoints. Each of the 10 activities contains interactions with multiple object classes in different environments. The dataset provides annotations regarding the activity and sub-activity labels, object labels, affordance labels and temporal segmentation of activities. The split of the related work~\cite{wu2021spatial} is used for a fair comparative evaluation.

\begin{figure}[t]
\centering
\includegraphics[width=.49\textwidth]{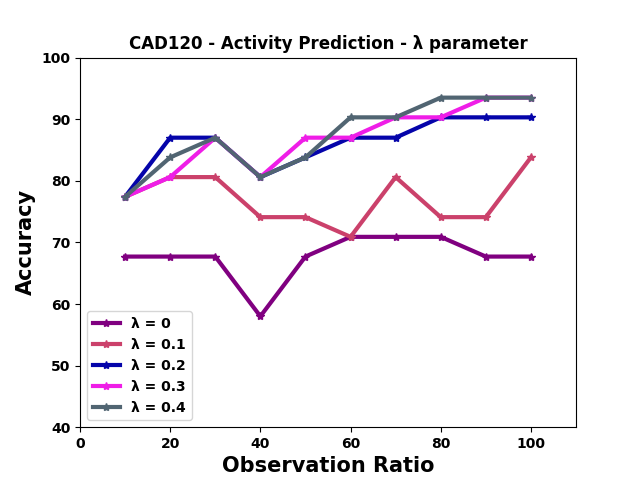} 
\includegraphics[width=.49\textwidth]{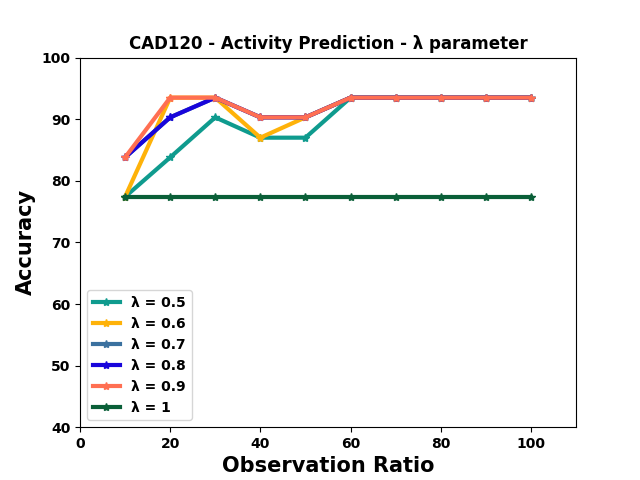} 
\caption{Exploration of the user-defined $\lambda$ parameter on the CAD-120 dataset. The values of the $\lambda$ parameter are in the range $[0, 1]$. Some curves may be partially visible due to occlusions. Plots are separated in two figures to aid readability.}
\label{fig:lamda}
\end{figure}

\subsection{Feature Extraction}
\label{ssec::feat_extraction}
The employed datasets are recorded from a third-person viewpoint, therefore they provide information for the whole or upper body of the acting subjects. We decided to align with the existing work of~\cite{manousaki2022segregational} and consider only the upper body human joints for both datasets.
For the MSR Daily Activity 3D Dataset the features used are the 3D joint angles and 3D skeletal joint positions~\cite{manousaki2022segregational}. Object classes and 2D object positions are obtained from YoloV4~\cite{bochkovskiy2020yolov4}. For the CAD-120 Dataset the 3D location of the joints of the upper body are used. As for the objects, the ground truth labels are used along with their 3D centroid locations~\cite{manousaki2021action,manousaki2022segregational}. 

\begin{figure}[t]
\centering
\includegraphics[width=.9\textwidth]{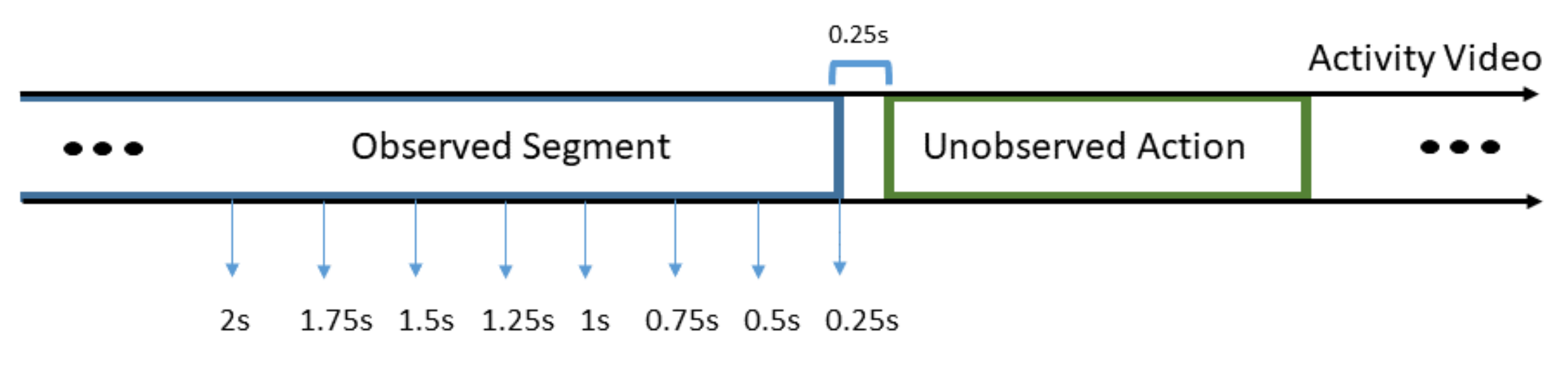} 
\caption{Observing the activity and making object predictions for [2s, 1.75s, 1.5s, 1.25s, 1s, 0.75s, 0.5s, 0.25s] before the beginning of the next action as in ~\cite{furnari2020rolling}.}
\label{fig:time}
\end{figure}

\subsection{Evaluation Metrics}
\label{ssec::measures}
\noindent {\textit{ Activity Prediction:}}
Activities are observed in a range from $10\%$ to $100\%$ of their total duration with steps equal to $10\%$. At every step, the accuracy of the predicted activity label is evaluated compared to the ground truth.

\vspace*{0.04cm}\noindent {\textit{ Next-Active-Object Prediction:}}
At variable time steps before the start of the next segment (see Fig.~\ref{fig:time}) where the next-active-object will be used, we estimate the accuracy of the predicted object label compared to the ground truth label. Also, we calculate the time at which the next-active-object will be used in the activity. For the aforementioned time steps the prediction error is calculated as the difference of the predicted time of use and the ground truth time, divided by the length of the video. 

\subsection{Results}
\label{ssec::results}
\noindent \textbf{Activity Prediction/Early Recognition:}
Activity label prediction is performed by considering observation ratios in chunks of $10\%$ until the end of the video. The label prediction at $100\%$ can be regarded as activity recognition. The test video is compared with all the reference videos by calculating the GED and is assigned to the label of the minimum. In Fig.~\ref{fig:msr} (left) a comparison of our method against the competitive methods for the MSR dataset is shown. Our method outperforms the works of Cao et al.~\cite{cao2020few}, Alfaifi et al.~\cite{alfaifi2020human} and others~\cite{panagiotakis2018graph,reily2018skeleton,alfaifi2020human} by a large margin. Our work also outperforms the method presented by Manousaki et al.~\cite{manousaki2022segregational} by a large margin at small observation ratios. Results of the competitive methods are taken as shown in ~\cite{manousaki2022segregational}.

CAD-120 is a challenging dataset due to the number of objects and their interchangeability in different executions of activities. In this dataset, our method outperforms the works of Manousaki et al.~\cite{manousaki2022segregational}, Furnari et al.~\cite{furnari2020rolling} and other competitive methods~\cite{kong2018action,zhou2018temporal,hu2018early} by a large margin. It also outperforms the approach of Wu et al.~\cite{wu2021spatial} that holds the state-of-art performance, for all observation ratios greater than $20\%$ (see Fig.~\ref{fig:msr}, right). The results of the~\cite{kong2018action,zhou2018temporal,hu2018early} and ~\cite{wu2021spatial} methods are taken from the work of Wu et al.~\cite{wu2021spatial} while for our previous work (Manousaki et al.~\cite{manousaki2022segregational}) we trained and tested using the activities (instead of actions) with the parameters mentioned in that paper.



\vspace*{0.04cm}\noindent {\textbf{The impact of parameter $\lambda$:}}
Edge weights are determined based on the proportion of the semantic and motion information they convey. This proportion is quantified by the user-defined parameter $\lambda$ (see Equation~(\ref{eq::weights})). In Fig.~\ref{fig:lamda} we present results that explore the impact of $\lambda$ on the performance of our approach on the CAD-120 dataset. When $\lambda = 0$ (only motion features) and $\lambda = 1$ (only semantic features) the results are alike in terms of having the lowest ability to make accurate predictions. Their combination carries a lot more information and gives the best results. Some values are not visible in the plots because for different values of the $\lambda$ parameter, accuracy values remain the same. After experimental evaluation the best value across datasets is $\lambda = 0.8$.


\begin{table}[t]
\centering
\begin{tabular}{||c|| c c c c c c c c c c c c c c c c ||} 
\hline
 &   \multicolumn{15}{c}{Next-Active-Object Prediction Accuracy}& \\
 \hline \hline
Time  & 2.00s && 1.75s && 1.50s && 1.25s&& 1.00s && 0.75s && 0.50s && 0.25s &\\
\hline
RULSTM~\cite{furnari2020rolling} &  18.6\% && 18.6\% && 18.0\% && 18.6\% && 18.6\% && 19.3\% && 20.0\%&& 22.0\% &  \\ 
\hline
GTF (Proposed) & 87.0\% && 87.0\% && 86.6\% && 89.1\% && 90.0\% && 91.0\% && 95.0\% && 97.0\% & \\ 
\hline
\end{tabular}
\caption{Next-active-object prediction accuracy for [2s, 1.75s, 1.5s, 1.25s, 1s, 0.75s, 0.5s, 0.25s] before the beginning of the next action for the CAD-120 dataset.}
\label{tbl::obj_acc}
\vspace{-0.9cm}
\end{table}

\vspace*{0.04cm}\noindent \textbf{Next-Active-Object Prediction:}
Our method is designed to accommodate videos captured from a third-person viewpoint as we need to have a view of the human joints and the surrounding objects. The most related work to ours is the work of Dessalene et al.~\cite{dessalene2021forecasting} which is currently limited only to egocentric videos. This does not allow for a comparison with that approach. We compare our method to the recent work of Furnari et al.~\cite{furnari2020rolling}. This work performs on both egocentric and third-view datasets and is the method that~\cite{dessalene2021forecasting} compares with. Their performance is comparable for the task of next-active-object prediction. However, instead of following their experimental scheme and evaluating only the accuracy of the prediction of the next-active-object, we also evaluate the accuracy of the prediction in relation to the time prior to the start of the action where the next-active-object will be used. Predictions are made in the range [2s, 1.75s, 1.5s, 1.25s, 1s, 0.75s, 0.5s, 0.25s] before the beginning of the action (see Fig.~\ref{fig:time}). As seen in Table~\ref{tbl::obj_acc} our method can correctly predict more objects as we move closer in time while~\cite{furnari2020rolling} can predict less accurately the objects and is not affected by the time horizon.
By comparing the graph of the partially observed video with those of the reference videos, the pair of graphs that have the smaller graph edit distance and object correspondences between the graphs are estimated (test and reference videos may have different number of objects). The work of Furnari et al.~\cite{furnari2020rolling} is tested using the CAD120 dataset and the publicly available implementation. We extracted the $1024$-dimensional features by using TSN~\cite{wang2016temporal} and calculated object features using the ground truth annotations. Their code accommodates the extraction of predictions at different seconds before the beginning of the action as described above.

\vspace*{0.04cm}\noindent \textbf{Next-Active-Object Time Prediction:}
Another aspect of great importance is the ability to forecast the time at which the object will be used in the activity. With the use of the GTF method we are able to compare the partially observed video with the reference videos from the training. After finding the pair of graphs that have the smaller graph edit distance, we acquire the information about object correspondences. This ability to infer the object correspondences between the two videos allows us to have the same number of objects between the videos in order to perform video alignment with the use of SSDTW. The alignment provides the ability to find the point of the reference video that corresponds to the current point in time in the test video (matching point). This projection of time from the reference video to the test one, permits the forecasting of the time at which the next-active-objects will be engaged in the interaction. The prediction error is calculated as the offset of the predicted time of use from the ground truth time of use of the next-active-object compared to the duration of the video. The error is calculated upon the correct predictions of the next-active-object. In Table~\ref{tbl::error} we observe that this error is low, which means that we are able to accurately predict the time at which the next-active-object will be used in the activity.



\begin{table}[t]
\centering
 \begin{tabular}{||c|| c c c c c c c c c c c c c c c c||} 
 \hline 
CAD120&  \multicolumn{15}{c}{Next-Active-Object Time Prediction Error} &\\
\hline\hline
Time  & 2.00s& & 1.75s& & 1.50s && 1.25s && 1.00s && 0.75s && 0.50s && 0.25s& \\
 \hline
GTF (Proposed) & 0.471 & & 0.463 && 0.46  && 0.457 && 0.443 && 0.405 && 0.36 && 0.325 &\\ 
 \hline
 \end{tabular}
\caption{Time prediction error is the offset of the predicted time of the next-active-object use to the ground truth time of use compared to video length. Predictions are made from 0.25s to 2s prior to the start of the next action.}
\label{tbl::error}
\vspace*{-0.4cm}
\end{table}

\begin{table}[t]
\centering
 \begin{tabular}{||c|| c c c c c c c c c ||} 
 \hline
CAD120 &   \multicolumn{8}{c}{Multiple Next-Active-Objects Prediction Accuracy} &\\
 \hline \hline
Observation Ratio & 10\% & 20\% & 30\% & 40\% & 50\% & 60\% & 70\% & 80\% & 90\%  \\ [0.5ex] 
 \hline
GTF (Proposed) &  41.7\% & 43.2\% & 45.6\% & 45.6\% & 47.1\% & 47.1\% & 48.6\% & 50\% & 55.9\%  \\ 
 \hline
 \end{tabular}
 \caption{Accuracy for predicting multiple next-active-objects for different observation ratios.   }
\label{tbl::mNaos}
\vspace{-0.55cm}
\end{table}

\vspace*{0.04cm}\noindent \textbf{Multiple Next-Active-Objects Prediction:}
Our method is capable of predicting not just one, but multiple next-active-objects. These predictions can be performed at different observation ratios from to $10\%$ to $90\%$ (an observation ratio equal to $100\%$ means that the whole video is observed, so next object prediction is not defined). The accuracy for each observation ratio for the predicted next-active-objects is presented at Table~\ref{tbl::mNaos}. The prediction is made through the correspondence of the objects between the reference and test graphs. By knowing the order in which the objects in the reference video are used, we can infer the order in which the objects of the test video will be used. After finding the matching point (see the previous section) we can infer the order of the matched objects from that point till the end. Prediction of multiple next-active-objects is challenging due to long time horizons involved and the related increased uncertainty. 

\vspace{-0.1cm}
\section{Conclusions}
\label{sec:conclusions}
We introduced GTF, a method that is based on matching complete and partially observed videos which are represented as graphs, with the use of Bipartite Graph matching. Human joints and objects were represented as nodes whereas their semantic and motion similarity was captured by the edges. We showed that through this formulation and process, we are able to perform activity and next-active-object prediction providing state-of-art results. Moreover, we proposed to solve the problem of predicting the time at which the next-active-object will be used as well as the prediction of multiple next-active-objects. 
Future research will be focused on compiling and experimenting with larger and more complex datasets of human-object interactions in which users will be handling a broader variety of objects in several ways. 
%
%
%
\section*{{Acknowledgements}}
This research was co-financed by Greece and the European Union (European Social Fund-ESF) through the Operational Programme ``Human Resources Development, Education and Lifelong Learning'' in the context of the Act ``Enhancing Human Resources Research Potential by undertaking a Doctoral Research'' Sub-action 2: IKY Scholarship Programme for PhD candidates in the Greek Universities. The research work was also supported by the Hellenic Foundation for Research and Innovation (HFRI) under the HFRI PhD Fellowship grant (Fellowship Number: 1592) and by HFRI under the ``1st Call for HFRI Research Projects to support Faculty members and Researchers and the procurement of high-cost research equipment'', project I.C.Humans, number 91.

\bibliographystyle{splncs04}
\bibliography{samplepaper}

\end{document}